\documentclass[11pt]{article}

\usepackage[preprint]{acl}
\usepackage{multirow}
\usepackage{makecell}
\usepackage{times}
\usepackage{latexsym}
\usepackage[T1]{fontenc}
\usepackage[utf8]{inputenc}
\usepackage{microtype}
\usepackage{inconsolata}
\usepackage{graphicx}
\usepackage{booktabs}
\usepackage{float}
\usepackage{amsmath}
\usepackage{enumitem}
\usepackage{tabularx}
\usepackage{algorithm}
\usepackage{algpseudocode}
\usepackage{bbm}
\usepackage{marvosym}
\title{\textsc{RealClawBench}: Live OpenClaw Benchmarks from Real Developer-Agent Sessions}
\author{
 \textbf{Zongwei Lv\textsuperscript{1$*$}},
 \textbf{Yaoming Li\textsuperscript{1$*$}},
 \textbf{Zhewen Tan\textsuperscript{1}},
 \textbf{Yilun Yao\textsuperscript{1}},
 \textbf{Yuxuan Tian\textsuperscript{1}},
 \\
 \textbf{Lin Sun\textsuperscript{2}},
 \textbf{Xiangzheng Zhang\textsuperscript{2}},
 \textbf{Weihong Lin\textsuperscript{2}},
 \textbf{Tong Yang\textsuperscript{1}},
 \textbf{Guangxiang Zhao\textsuperscript{2\Letter}}
\\
 \textsuperscript{1}Peking University,
 \textsuperscript{2}Qiyuan Tech
\\
 \small{
    \textsuperscript{\Letter}\textbf{Correspondence:}
   \href{mailto:zhaoguangxiang@pku.edu.cn}{zhaoguangxiang@pku.edu.cn}
 }
 \\
 {\normalsize \hspace{0.25em}\textbf{Project Repository:} \url{https://realclawbench.github.io/index.html}}
}
\begin{document}
\maketitle
\let\thefootnote\relax\footnotetext{
    \textsuperscript{$*$}Equal contribution \
    \textsuperscript{\Letter} Correspondence
}

\begin{abstract}
Agent benchmarks should reflect what users actually ask deployed agents to do, yet existing benchmarks often miss key realism properties of real developer-agent sessions. We introduce \textsc{RealClawBench}, a live benchmark framework built from real OpenClaw sessions to capture the distribution, diversity, and real-world difficulty of deployed agent use. Real user requests are challenging to benchmark because they often depend on local execution environments, involve implicit or underspecified intent, and require nontrivial verification. \textsc{RealClawBench} addresses these challenges with two core mechanisms: reconstructed execution environments and deterministic verifiable scorers, which together convert real sessions into reproducible, automatically scored tasks. The resulting release contains 281 executable tasks sampled from a much larger real-session pool while preserving the source distribution, with maximum final-vs-source Jensen-Shannon divergence of 0.0448. Evaluating 14 contemporary models shows that the best system solves only 65.8\% of tasks, revealing substantial headroom on realistic developer-agent workloads. By turning real deployed sessions into controlled evaluation instances, \textsc{RealClawBench} provides a practical path toward benchmarks that better measure agent capability in actual use. 
\end{abstract}

\section{Introduction}
\begin{figure}[t]
\centering
\includegraphics[width=0.98\columnwidth]{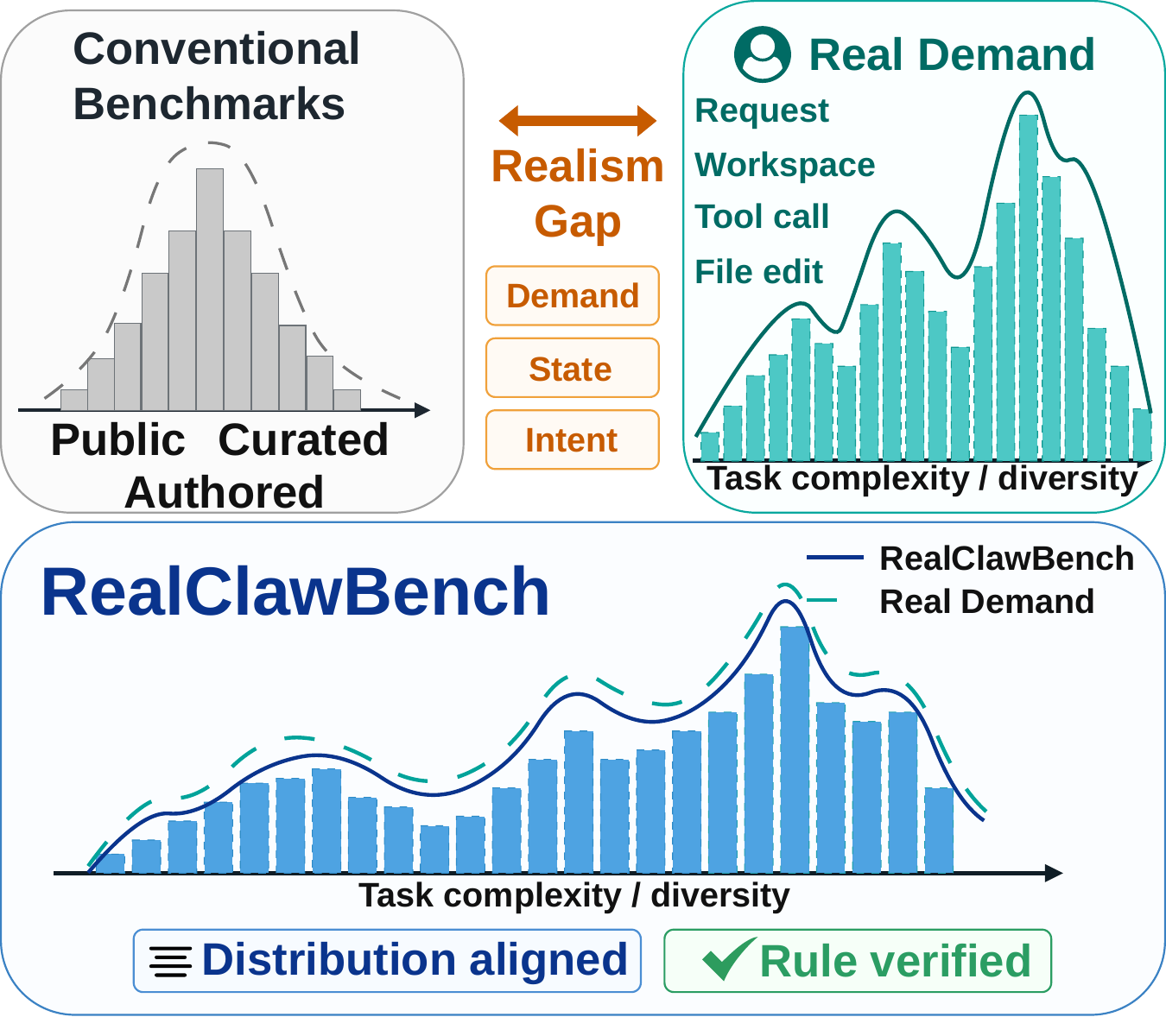}
\caption{Conceptual overview of the realism gap and the \textsc{RealClawBench} response.}
\label{fig:realism-gap}
\end{figure}
\begin{figure*}[t]
\centering
\includegraphics[width=0.98\textwidth]{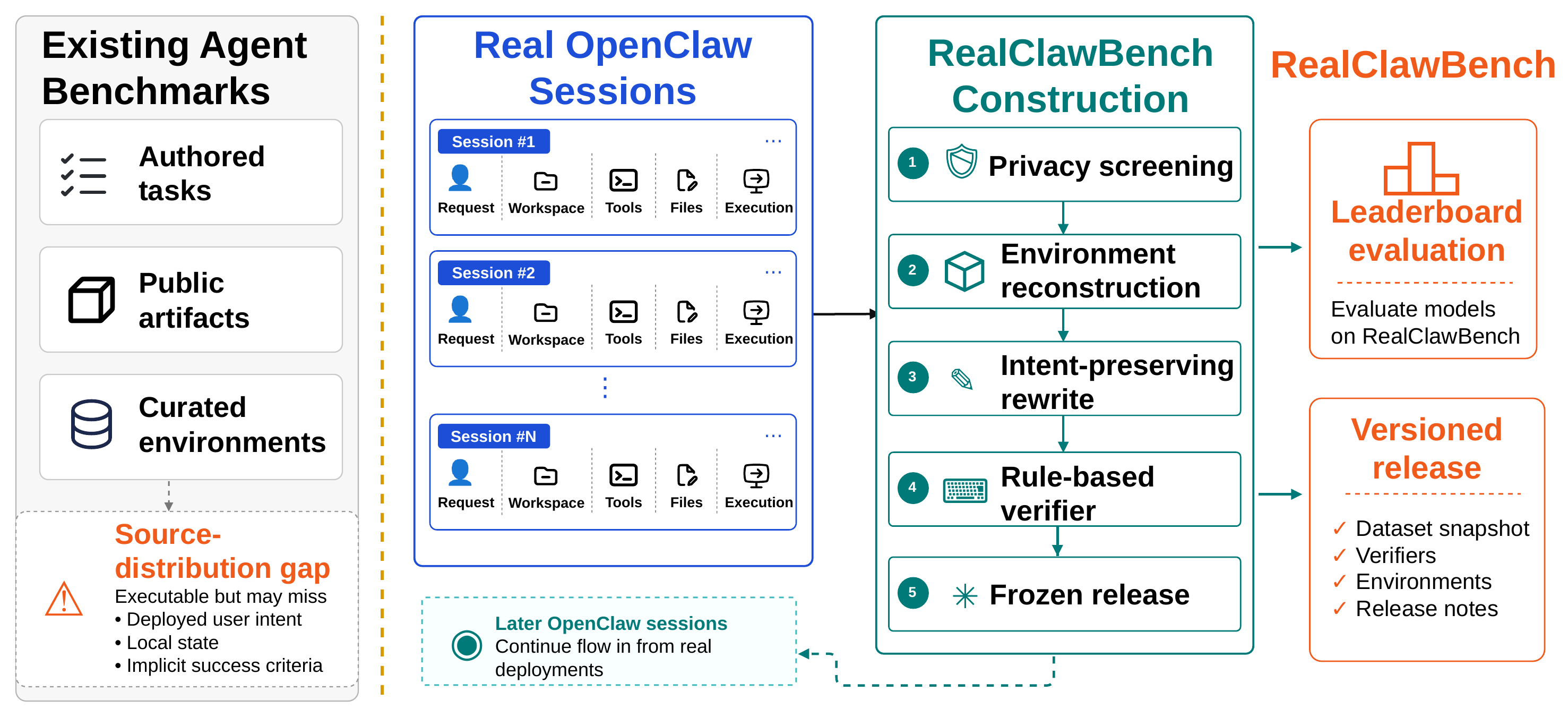}
\caption{\textsc{RealClawBench} pipeline from real OpenClaw sessions to frozen releases and live updates. Real sessions define the source distribution, while filtering, sampling, environment reconstruction, request rewriting, and rule-based verification turn that stream into controlled evaluation artifacts.}
\label{fig:pipeline}
\end{figure*}
Developer-facing agents are becoming deployed work systems rather than only chat interfaces. The most representative one is OpenClaw\footnote{OpenClaw: \url{https://github.com/openclaw/openclaw}}, which has attracted substantial attention due to its strong practical value, becoming one of the fastest-growing projects on GitHub in terms of stars. In OpenClaw, users ask agents to inspect repositories, call tools, edit files, run commands, and carry out multi-step workflows inside stateful workspaces. This shift changes what evaluation should measure: success should mean completing the user's intended task in the environment where that task arose. Prior work has made agent evaluation more tool-aware, executable, and time-sensitive. ReAct-style agents interleave reasoning with actions \citep{yao2022react}, Toolformer studies API use \citep{schick2023toolformer}, and WebGPT shows evidence gathering in interactive environments \citep{nakano2021webgpt}. SWE-bench evaluates repository-level issue resolution \citep{jimenez2024swe}, WebArena studies long-horizon web tasks \citep{zhou2024webarena}, AgentBench and GAIA broaden evaluation across tools and domains \citep{liu2024agentbench,mialon2024gaia}, and LiveBench/LiveCodeBench update public-source tasks to reduce staleness \citep{white2024livebench,jain2025livecodebench}. NL2Repo-Bench \citep{ding2025nl2repo} studies long-horizon repository generation from natural-language requirements. However, these benchmarks still largely begin from authored tasks, public artifacts, or curated environments rather than deployed developer-agent sessions. As Figure~\ref{fig:realism-gap} shows, this creates a realism gap: a benchmark can be executable and difficult while still smoothing away real demand, local workspace state, implicit user intent, and naturally occurring success criteria.

\textbf{Building a realistic benchmark from deployed sessions is challenging because each task must preserve a rich, recoverable execution environment and support deterministic verification.} Real sessions may contain private context, depend on local project state, express intent implicitly, or lack deterministic success criteria. \textsc{RealClawBench} addresses this challenge by treating real OpenClaw sessions as the source distribution for benchmark construction \citep{openclaw2026runtime}. As Figure~\ref{fig:pipeline} shows, the pipeline samples deployed sessions, filters low-quality or unsafe cases, reconstructs execution environments, rewrites requests into standalone instructions, and builds deterministic verifiers. The resulting benchmark therefore measures agents on tasks closer to what users actually experience in deployment. Because the same construction process can be rerun on later sessions, \textsc{RealClawBench} also supports live releases that reduce staleness and data contamination. This paper makes the following contributions:

\begin{itemize}[leftmargin=*]
    \item \textbf{Realism gap:} We identify the realism gap between executable agent benchmarks and deployed developer-agent use.
    \item \textbf{Session-to-task pipeline:} We present a pipeline that converts real sessions into reproducible, privacy-screened tasks with reconstructed environments and deterministic verifiers.
    \item \textbf{Live benchmark release:} We release \textsc{RealClawBench}, a live, distribution-anchored benchmark that reveals substantial headroom for current agents on realistic workloads.
\end{itemize}

\section{Related Work}

\paragraph{Static and live benchmarks}
MMLU \citep{hendrycks2020measuring}, BIG-bench \citep{srivastava2023beyond}, and HELM \citep{liang2022holistic} established broad standardized evaluations for language-model knowledge, reasoning, and reporting discipline.
LiveBench \citep{white2024livebench} and LiveCodeBench \citep{jain2025livecodebench} improve temporal validity by updating items from recent public sources.
These benchmarks make model comparison more systematic and less stale, but their items are collected or authored outside deployed agent use. They therefore do not directly measure whether an agent can solve tasks from real sessions with local state, tool traces, and implicit user intent.

\begin{table*}[t]
\centering
\small
\begin{tabular}{lrrrrrrrrr}
\toprule
Benchmark
& \multicolumn{4}{c}{Distribution}
& \multicolumn{5}{c}{User-intent signals} \\
\cmidrule(lr){2-5}\cmidrule(lr){6-10}
& JSD$_{\text{base}}\downarrow$
& TV$_{\text{base}}\downarrow$
& JSD$_{\text{live}}\downarrow$
& TV$_{\text{live}}\downarrow$
& Intent$\uparrow$
& Local$\uparrow$
& Artifact$\uparrow$
& Constr.$\uparrow$
& Exec.$\uparrow$ \\
\midrule
RealClawBench & \textbf{0.146} & \textbf{0.271} & \textbf{0.120} & \textbf{0.263} & \textbf{6.37/8} & \textbf{38.4\%} & \textbf{96.8\%} & \textbf{91.8\%} & \textbf{75.4\%} \\
Claw-Eval & 0.335 & 0.538 & 0.402 & 0.621 & 3.20/8 & 0.3\% & 39.7\% & 19.3\% & 24.0\% \\
WildClawBench & 0.283 & 0.370 & 0.336 & 0.427 & -- & -- & -- & -- & -- \\
SWE-bench & 0.615 & 0.804 & 0.640 & 0.823 & 3.85/8 & 16.0\% & 42.3\% & 42.3\% & 69.3\% \\
WebArena & 1.000 & 1.000 & 1.000 & 1.000 & 0.38/8 & 0.0\% & 0.0\% & 0.0\% & 1.4\% \\
Terminal-Bench 2.0 & 0.905 & 0.971 & 0.922 & 0.978 & 4.73/8 & 3.4\% & 51.7\% & 79.8\% & 58.4\% \\
WorkArena & 0.607 & 0.798 & 0.710 & 0.871 & -- & -- & -- & -- & -- \\
OSWorld & 1.000 & 1.000 & 1.000 & 1.000 & -- & -- & -- & -- & -- \\
AgentBench-OS & -- & -- & -- & -- & 2.28/8 & 9.7\% & 0.0\% & 41.7\% & 23.6\% \\
GAIA & -- & -- & -- & -- & 0.95/8 & 0.0\% & 0.0\% & 4.5\% & 18.2\% \\
\bottomrule
\end{tabular}
\caption{\textbf{Alignment with deployed OpenClaw demand.} Distances are measured against two disjoint references: $W_{\text{base}}$, the 76,155 construction-source tool-use sessions, and $W_{\text{live}}$, the 11,638 sessions of a later window that contributed no released task. Lower JSD and TV indicate a closer match to real OpenClaw demand. Text-level user-intent signals are computed from public prompt/task text using transparent rules. RealClawBench is closest under both references and retains substantially more local, artifact-oriented, constrained, environment-dependent user-intent signals. Dashes indicate that the public artifact does not support a fair computation for that metric. The detailed computation procedure for the metrics in this table is provided in Appendix~\ref{app:realness-metrics}.}
\label{tab:realism-gap}
\end{table*}

\paragraph{Agent benchmarks}
Agent benchmarks move beyond static question answering by evaluating planning, tool use, and stateful execution. API-Bank \citep{li2023apibank} and StableToolBench \citep{guo2025stabletoolbench} evaluate tool-augmented models and tool-learning stability, while $\tau$-bench adds multi-turn tool-agent-user interaction under structured domain rules \citep{yao2024taubench}.
AgentBench \citep{liu2024agentbench}, WebArena \citep{zhou2024webarena}, GAIA \citep{mialon2024gaia}, and SWE-bench \citep{jimenez2024swe} place agents in tool, web, and software environments with executable feedback.
OSWorld \citep{xie2024osworld} and WorkArena \citep{drouin2024workarena} extend the setting to computer-use and enterprise web tasks, and Terminal-Bench evaluates agents on hard, realistic command-line tasks \citep{merrill2026terminalbenchbenchmarkingagentshard}.
SWE-bench Multimodal \citep{yang2024swebenchmultimodal} and MLE-bench \citep{chan2025mlebench} broaden developer-agent evaluation to visual software issues and machine-learning engineering.
These benchmarks reveal failures that static tests miss, but their sources remain largely authored, public-artifact-derived, simulated, or curated.
Our focus is complementary: we ask whether the benchmark's source distribution matches deployed developer-agent demand.

\paragraph{Real-user data}
Real-user datasets show why source distributions matter. LMSYS-Chat-1M \citep{zheng2024lmsys} and WildChat \citep{zhao2024wildchat} capture large-scale user conversations in the wild.
Chatbot Arena \citep{chiang2024chatbotarena} and MT-Bench \citep{zheng2023judging} popularized preference-based and judge-based evaluation, while WildBench builds challenging tasks from real-user queries \citep{lin2024wildbench,li2026primerposttrainingreasoningdata}.
These resources capture demand and preference signals that authored prompts often miss.
However, they are primarily conversational or output-oriented: they generally do not reconstruct local workspaces, expose the tool-use environment, or verify the final state after an agent acts.

\paragraph{OpenClaw evaluations}
OpenClaw provides an agent runtime with tools, workspaces, and harness-level execution surfaces \citep{openclaw2026runtime}.
WildClawBench \citep{ding2026wildclawbench}, Claw-Eval \citep{ye2026clawevaltrustworthyevaluationautonomous}, and ClawBench \citep{openclaw2026clawbench} cover OpenClaw-native, full-stack, or trustworthy autonomous-agent settings. Harness-centered evaluation shows that execution-layer design can affect agent behavior in realistic workflows \citep{yao2026harness}. OpenClaw safety benchmarks study persistent-state attacks \citep{wang2026agentasset} and trajectory-level risk \citep{yang2026atbenchclaw}.

Together, these studies show that runtime matters for agent evaluation.
\textsc{RealClawBench} is complementary: it uses the same deployed ecosystem as its starting point, but asks a different question by converting actual OpenClaw user sessions into privacy-screened, executable, and automatically scored benchmark tasks.

\section{The Realism Gap in Agent Benchmarks}
\label{sec:realism-gap}


Executable agent benchmarks have made evaluation more operational, but executability alone does not guarantee realism. A benchmark can require tool use, environment interaction, and executable feedback while still measuring a distribution that differs from deployed agent use. We therefore compare existing agent benchmarks with the OpenClaw tool-use stream of $76{,}155$ deployed sessions. Table~\ref{tab:realism-gap} measures this gap from two views: distributional distance from deployed task categories, and user-intent signals tied to local artifacts, explicit constraints, and environment-dependent state.

\paragraph{Intent signals.}
The intent columns of Table~\ref{tab:realism-gap} are computed by eight deterministic binary functions evaluated on the public instruction alone; rubrics, reference solutions, workspaces, verifiers, and metadata are never read. For instruction $x$ and a fixed multilingual pattern set $\mathcal{R}_j$, we set $f_j(x)=\mathbbm{1}[\exists r\in\mathcal{R}_j:\ r\ \text{matches}\ x]$, so repeated matches within a dimension add no score. Each $f_j$ is active iff the instruction:
\begin{enumerate}[label=$f_{\arabic*}$,labelsep=0.5em,leftmargin=2.1em,topsep=3pt,itemsep=2pt,parsep=0pt]
\item \emph{Request}: asks to create, fix, inspect, extract, run, or modify.
\item \emph{Workspace}: names an existing workspace, repository, directory, path, or project.
\item \emph{Artifact}: requires a concrete artifact to be read, produced, modified, or returned.
\item \emph{Constraint}: specifies a path, format, version, quantity, prohibition, or resource limit.
\item \emph{Procedure}: gives at least two enumerated or sequenced operations.
\item \emph{Execution}: requires running a command, test, build, query, installation, or conversion.
\item \emph{State}: depends on pre-existing workspace, filesystem, conversation, or execution state.
\item \emph{Completion}: gives an observable stopping condition, such as a required path, passing tests, or expected output.
\end{enumerate}
We summarize a benchmark $B$ by its per-signal shares and their total:
\begin{align*}
\mathrm{Share}_j(B) &= |B|^{-1}\sum_{x\in B} f_j(x),\\
\mathrm{Intent}(B) &= \sum_{j=1}^{8}\mathrm{Share}_j(B)\in[0,8].
\end{align*}
The four reported component columns are $\mathrm{Share}_2$ (Local), $\mathrm{Share}_3$ (Artifact), $\mathrm{Share}_4$ (Constr.), and $\mathrm{Share}_6$ (Exec.). These signals measure explicit grounding in instruction text, not latent intent, difficulty, or universal realism. Appendix~\ref{app:realness-metrics} releases the complete patterns and mappings.

\paragraph{Reference windows.} Because \textsc{RealClawBench} is built from $W_{\text{base}}$, we also recompute every distance against $W_{\text{live}}$, a later, record-disjoint window that contributed no released task, with the taxonomy and mappings held fixed. \textsc{RealClawBench} remains closest under this held-out reference, so the ordering does not follow from matching released tasks to their own source records. Values of $1.000$ indicate disjoint support under an OpenClaw-scoped taxonomy, not that such tasks are unrealistic; the comparison measures alignment with one deployed population, not universal realism.

The comparison shows that existing benchmarks remain separated from deployed developer-agent use in both distribution and intent signals. Many benchmarks are executable, but they often underrepresent the local, artifact-centered, and state-dependent requests that appear in real sessions. This supports the central premise of \textsc{RealClawBench}: realistic agent evaluation should not only ask whether an agent can solve controlled tasks, but whether it can solve tasks from the distribution and context of deployed agent use.

\section{From Real Sessions to Verifiable Tasks}
\label{sec:construction}
The source data for \textsc{RealClawBench} come from production OpenClaw usage logs, where real users interact with a deployed developer-agent system for repository inspection, file editing, command execution, data extraction, and project-building workflows.
These logs are not researcher-authored prompts: they are user-agent sessions produced during ordinary OpenClaw use, containing user requests, conversation context, available tools, tool calls, workspace traces, and observed outcomes.
\textsc{RealClawBench} converts these deployed sessions into reproducible benchmark tasks. Raw sessions cannot be used directly because they may
contain private context, depend on local workspace state, express intent
implicitly, or lack deterministic success criteria. Each released instance is constructed through the workflow in Figure~\ref{fig:pipeline}:
\[
b=(x,E_0,V,m),
\]
where $x$ is a standalone instruction, $E_0$ a sanitized initial workspace, $V$ an automatic verifier, and $m$ stratification metadata. We keep only sessions with recoverable intent, reconstructable state, auditable verification, and release-safe content.

\subsection{Anchoring the Source Distribution}
\label{sec:source-anchoring}

Using this reference stream, the first stage defines the empirical distribution that the benchmark should preserve. We remove framework-level noise, deduplicate repeated retries of the same task, and retain tool-mediated developer-agent requests. The resulting tool-use stream captures the mix of
files, commands, tools, workspace state, and user intents seen in deployment.
\begin{table}[t]
\centering
\small
\begin{tabular}{llrr}
\toprule
Stage & Name & $n$ & Drop \\
\midrule
$N_0^A$ & Raw calls & 450,766 & -- \\
$N_0^S$ & Sessions & 110,170 & 75.6\% \\
$N_0^T$ & Tool-use & 76,155 & 30.9\% \\
$N_1$ & Cleaned & 40,713 & 46.5\% \\
$N_2$ & HQ pool & 6,995 & 82.8\% \\
$N_3$ & Scorable & 5,260 & 24.8\% \\
$C$ & Candidates & 414 & 92.1\% \\
$F$ & Final & 281 & 32.1\% \\
\bottomrule
\end{tabular}
\caption{Construction funnel from raw API calls to the released evaluation set. Drop percentages are relative to the previous stage, showing where session folding, tool-use filtering, quality screening, scorer construction, and final review reduce the pool.}
\label{tab:exp1-funnel}
\end{table}
Importantly, sampling is anchored before final quality filtering. Candidate seeds are drawn from the cleaned tool-use stream rather than from the final high-quality pool, so later filtering cannot silently redefine the target distribution. Common categories follow their empirical mass, while rare categories receive a minimum diagnostic allocation to support fine-grained analysis. Table~\ref{tab:exp1-funnel} summarizes the resulting construction funnel, from raw calls to the released evaluation set.

\subsection{Projecting Sessions into Benchmark Instances}
\label{sec:projection}

\begin{figure}[t]
\centering
\includegraphics[width=0.98\columnwidth]{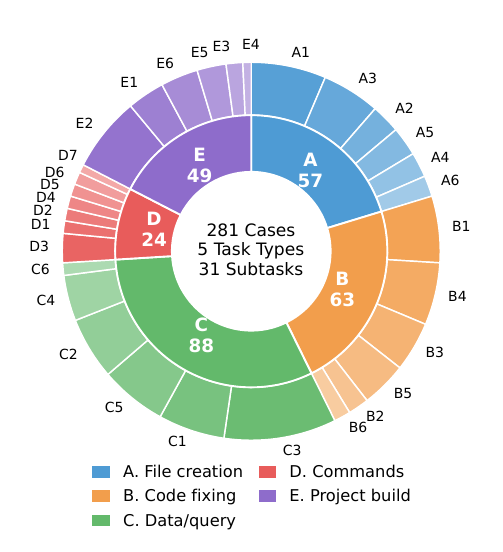}
\caption{Task composition of the final evaluation set. The inner ring shows the top-level task types, and the outer ring shows subtasks. Full task and subtask names are listed in Appendix~\ref{app:task-subtask-distribution}.}
\label{fig:exp1-task-distribution}
\end{figure}

For each sampled seed, we apply three coupled transformations. First,
environment reconstruction materializes the initial workspace \(E_0\) from
session evidence, including files, dependencies, command context, schemas, and expected output locations. Tasks that depend on private
services are fixture-backed when possible and rejected otherwise. Second,
request rewriting turns the original user message into a standalone
instruction \(x\), resolving local references and conversational ellipsis while
removing private identifiers. Rewrites must preserve the task category and
must not add solution hints or unavailable evidence. Third, verifier
construction builds a deterministic program \(V\) over the final workspace,
stdout, and bounded subprocesses. Verifiers check structural outcomes such as
file existence, schema validity or content matches. Tasks requiring subjective or aesthetic judgment are
removed.
Figure~\ref{fig:exp1-task-distribution} shows this composition: the inner ring gives the task types, and the outer ring gives the subtasks used for fine-grained analysis.

\subsection{Fidelity Review and Release Filtering}
\label{sec:fidelity-review}

Draft instances are probed under the leaderboard harness to find benchmark
bugs, not to score models. We repair missing context, nondeterminism,
ambiguous wording, leaky instructions, over-broad verifiers, and
environment-free shortcuts only when supported by the original session
evidence. Otherwise the instance is rejected. Released tasks must pass
reconstructability, rewrite-preservation, verifier-validity, and probing
checks. We release only sanitized instructions, workspaces, verifiers,
rubrics, metadata, and manifests. Raw logs and private or unreleasable
artifacts are excluded.
\section{Benchmark Protocol}
\label{sec:protocol}

An evaluated agent receives the standalone instruction \(x\) and access to
the reconstructed workspace \(E_0\) through a shared OpenClaw harness. For
each instance, the harness restores a clean workspace, exposes fixed tools and
resource budgets, runs the agent under the same context, turn, timeout, and
final-answer protocol, records the trajectory, and invokes the verifier \(V\)
on the final state.

For model \(A\), the sample-average verifier pass rate is the main leaderboard metric.
We also define a distribution-weighted score
\begin{equation}
    S_t(A) = \sum_c \hat{p}_t(c) \cdot
    \frac{1}{|B_{t,c}|}\sum_{b \in B_{t,c}} V_b(A(b))
\end{equation}
where \(B_{t,c}\) is the set of released instances in category \(c\) and
\(\hat{p}_t(c)\) is the estimated deployed-session mass of category \(c\).
Because deployed distributions are imbalanced, the sample score estimates performance on the released set.
The weighted score estimates observed-demand performance, while category and subtask scores expose failures on rarer workflows.
We also report cost and tool-use statistics. Leaderboard scores are computed only from
deterministic verifiers, not LLM judges.

Each release is frozen and versioned by identifier. Live releases reuse the same construction map $T$ on a later, non-overlapping OpenClaw window and are published separately. Release cards record the log window, filtering counts, sampling weights, category distributions, verifier types, privacy review status, limitations, and distribution drift. Released artifacts include standalone instructions, sanitized workspaces, verifiers, rubrics, metadata, and an evaluation manifest. Raw sessions, private identifiers, credentials, proprietary files, and unreleasable service traces are excluded.
\begin{figure*}[t]
\centering
\includegraphics[width=0.95\textwidth]{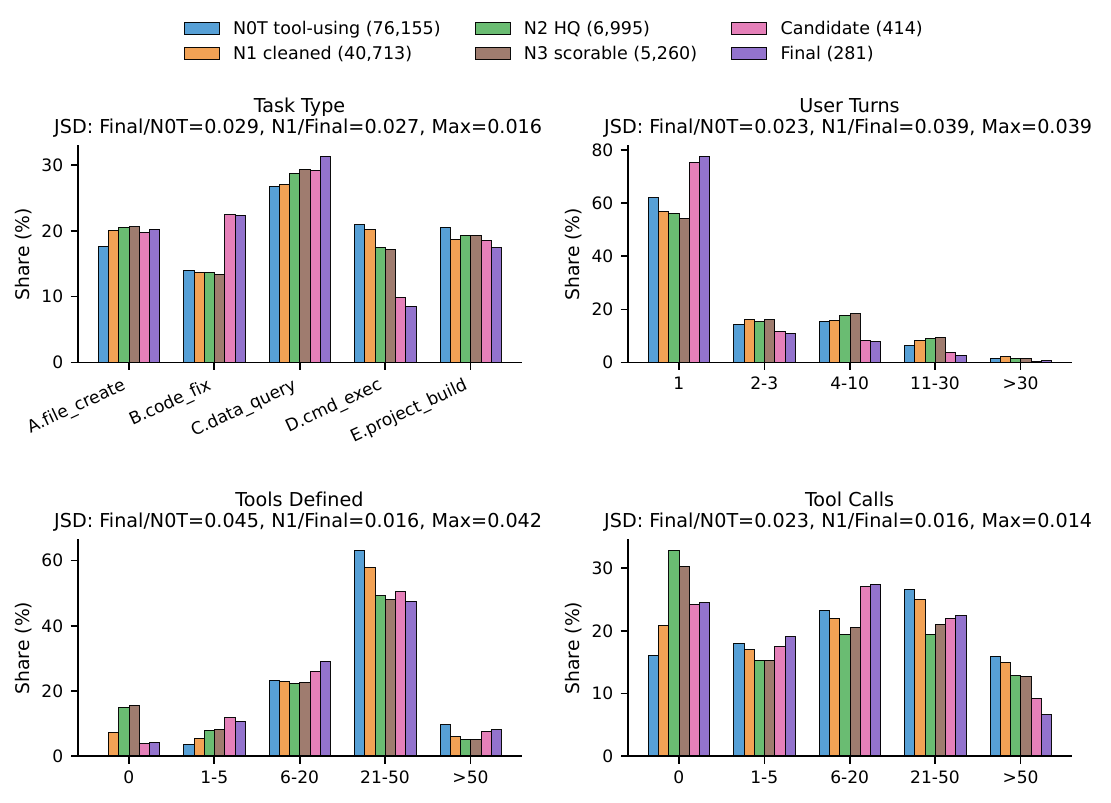}
\caption{Construction-stage distribution fidelity. Across task type, user turns, tools defined, and tool calls, the final 281-task release remains close to the 76,155-session tool-using reference stream, with final-vs-reference JSD below 0.05 bits on all axes. This shows that filtering, scorer construction, and final review preserve the observable shape of real OpenClaw usage rather than introducing substantial distribution drift.}
\label{fig:exp1-distribution}
\end{figure*}
\section{Experiments and Results}

We evaluate \textsc{RealClawBench} along three questions: whether the construction pipeline preserves the distribution of real OpenClaw tool-use sessions, whether the resulting tasks distinguish contemporary agents under a shared harness, and whether the same protocol can be reused on later logs for live, versioned releases. Unless otherwise noted, all evaluations use the 281-case release in Table~\ref{tab:exp1-funnel} and Figure~\ref{fig:exp1-task-distribution}. We run each case with the OpenClaw harness described in Section~\ref{sec:protocol}: the harness restores the reconstructed workspace, exposes the same file and command tools, records the trajectory, applies the task-specific deterministic verifier, and enforces a 600-second timeout. The headline metric is case-level verifier pass rate. We also report task-type and subtask macro averages to expose long-tail behavior under the imbalanced real-user distribution. Pass@3 denotes the fraction of cases solved in at least one of three runs. Runtime, cost, tool calls, tool failures, and timeouts are diagnostic metrics. All leaderboard scores come from deterministic verifiers. The LLM auditor in Appendix~\ref{app:scorer-validation} is used only for scorer analysis, never for model ranking.

\begin{table}[t]
\centering
\small
\begin{tabular}{lrrrr}
\toprule
Comp. & Task & Turns & Tools & Calls \\
\midrule
$F/N_0^T$ & 0.0294 & 0.0232 & 0.0448 & 0.0233 \\
$N_0^T/N_1$ & 0.0009 & 0.0028 & 0.0419 & 0.0027 \\
$N_1/N_2$ & 0.0010 & 0.0009 & 0.0143 & 0.0138 \\
$N_2/N_3$ & 0.0001 & 0.0002 & 0.0001 & 0.0007 \\
$N_3/C$ & 0.0155 & 0.0392 & 0.0326 & 0.0081 \\
$C/F$ & 0.0008 & 0.0010 & 0.0012 & 0.0019 \\
$N_1/F$ & 0.0265 & 0.0389 & 0.0161 & 0.0162 \\
\bottomrule
\end{tabular}
\caption{Jensen-Shannon divergence in bits across construction stages and observable axes. The table tests whether filtering and review distort the reference distribution. All final-vs-reference and adjacent-stage shifts remain below 0.05 bits, indicating that aggressive filtering preserves the observable source distribution.}
\label{tab:exp1-jsd}
\end{table}
\subsection{Construction Fidelity}

This experiment asks whether \textsc{RealClawBench} can filter aggressively without drifting away from the measured user distribution. The audit uses the construction funnel in Table~\ref{tab:exp1-funnel}. We use $N_0^T$ as the primary reference because the benchmark targets tool-mediated tasks. The full session pool is kept only as a boundary check, since it includes many non-agentic requests. The role of each filtering stage is given in Appendix~\ref{app:construction-funnel}. Here we evaluate whether filtering and review preserve the measured shape of $N_0^T$. We measure filtering-induced shift against $N_0^T$ along four observable axes: task type, real user turns, tools defined, and tool calls, using Jensen-Shannon divergence (JSD) in bits. Lower values indicate closer distributions.

\newcommand{\first}[1]{\textbf{#1}}
\newcommand{\second}[1]{\underline{#1}}
\newcommand{\third}[1]{\textit{#1}}
\begin{table*}[t]
\centering
\small
\setlength{\tabcolsep}{3pt}
\begin{tabular}{l l ccccccccc}
\toprule
\multirowcell{2}{Rank} & \multirowcell{2}{Model} & \multirowcell{2}{Sample} & \multicolumn{6}{c}{Task} & \multirowcell{2}{Subtask} & \multirowcell{2}{pass@3} \\
\cmidrule(lr){4-9}
& & & \makecell{file\\creation} & \makecell{code\\fixing} & \makecell{data/codebase\\querying} & \makecell{command\\execution} & \makecell{project\\building} & \makecell{Avg.} & & \\
\midrule
1 & Claude Opus 4.7 & \first{65.8\%} & \second{66.1\%} & \third{58.7\%} & \first{68.6\%} & \second{65.3\%} & \first{70.1\%} & \first{65.7\%} & \first{65.9\%} & \second{75.1\%} \\
2 & GPT-5.5 & \second{65.0\%} & \first{70.2\%} & 57.7\% & \second{64.4\%} & \second{65.3\%} & \second{69.4\%} & \second{65.4\%} & \second{65.8\%} & \second{75.1\%} \\
3 & MiMo V2.5 Pro & \third{60.1\%} & 54.4\% & \first{61.4\%} & 59.8\% & \third{63.9\%} & \third{63.9\%} & \third{60.7\%} & \third{63.0\%} & \third{74.7\%} \\
4 & DeepSeek V4 Pro & 59.8\% & 52.6\% & \second{60.3\%} & \third{61.7\%} & \third{63.9\%} & 61.9\% & 60.1\% & 61.5\% & \first{76.2\%} \\
5 & GLM 5.1 & 57.4\% & 53.8\% & 51.9\% & 58.7\% & \first{66.7\%} & 61.9\% & 58.6\% & 60.3\% & 70.8\% \\
6 & Kimi K2.6 & 57.7\% & 59.1\% & 56.1\% & 58.3\% & \first{66.7\%} & 52.4\% & 58.5\% & 60.2\% & 73.0\% \\
7 & Gemini 3.1 Pro & 57.4\% & 55.0\% & 52.4\% & 60.2\% & 59.7\% & 60.5\% & 57.6\% & 58.9\% & 69.0\% \\
8 & DeepSeek V4 Flash & 56.0\% & 60.2\% & 50.3\% & 55.7\% & 62.5\% & 55.8\% & 56.9\% & 58.1\% & 71.2\% \\
9 & Qwen 3.6 Plus & 55.8\% & \third{61.4\%} & 47.6\% & 56.4\% & 54.2\% & 59.2\% & 55.8\% & 56.1\% & 67.6\% \\
10 & Claude Sonnet 4.6 & 55.3\% & 56.1\% & 49.7\% & 59.1\% & 52.8\% & 55.8\% & 54.7\% & 55.5\% & 70.5\% \\
11 & MiniMax M2.7 & 53.7\% & 53.8\% & 49.2\% & 57.2\% & 55.6\% & 52.4\% & 53.6\% & 53.7\% & 68.3\% \\
12 & Claude Opus 4.6 & 52.1\% & 46.8\% & 50.3\% & 51.5\% & 58.3\% & 58.5\% & 53.1\% & 52.6\% & 67.3\% \\
13 & Gemma 4 31B & 45.7\% & 25.1\% & 47.6\% & 56.1\% & 52.8\% & 44.9\% & 45.3\% & 47.5\% & 51.6\% \\
14 & GPT-OSS 120B & 42.7\% & 35.7\% & 47.6\% & 42.4\% & 56.9\% & 38.1\% & 44.2\% & 43.1\% & 57.7\% \\
\bottomrule
\end{tabular}
\caption{Main leaderboard on the \textsc{RealClawBench} evaluation set, including average scores, task-type scores, subtask macro-average, and pass@3. Under Task, Avg.\ is the macro-average over the five task types. First, second, and third places in each metric column are marked with bold, underline, and italics, respectively.}
\label{tab:exp2-leaderboard}
\end{table*}
In Table~\ref{tab:exp1-jsd}, A/B denotes the comparison between stages A and B, $C$ denotes candidate case cards, and $F$ denotes the final set. For tool-call comparisons, unrecoverable trajectory bins are excluded on both sides. Upstream deterministic labels are mapped into the top-level task types, and the final cases remain traceable to OpenClaw sessions. Appendix~\ref{app:task-subtask-distribution} gives the full task table, and Appendix~\ref{app:construction-funnel} gives the funnel notes.
Figure~\ref{fig:exp1-distribution} reports final-vs-reference JSD, direct $N_1$-to-final JSD, and the largest adjacent-stage drift in its panel subtitles. Table~\ref{tab:exp1-jsd} and Figure~\ref{fig:exp1-distribution} show that final-vs-reference JSD stays below 0.05 bits on all four axes: 0.0294 for task type, 0.0232 for user turns, 0.0448 for tools defined, and 0.0233 for tool-call count.
All adjacent-stage comparisons are also below 0.05 bits, including candidate-to-final manual review, and the direct $N_1$-to-final comparison remains below 0.039 bits on every axis.
Thus \textsc{RealClawBench} remains selective without losing the measured shape of $N_0^T$.

\subsection{Main Leaderboard}

We next test whether the benchmark separates current agents under a shared execution protocol. Table~\ref{tab:exp2-leaderboard} reports the 14-model leaderboard on the released benchmark. The evaluated systems span proprietary frontier APIs and open-weight or openly documented models from major model families. They include Claude Opus/Sonnet variants \citep{anthropic2026claudeoverview}, OpenAI GPT-5.5 \citep{openai2026gpt55} and GPT-OSS \citep{openai2025gptoss}, Google Gemini \citep{google2026gemini31} and Gemma \citep{google2026gemma4}, and MiMo \citep{xiaomi2026mimo25}, DeepSeek \citep{deepseek2026v4}, GLM \citep{zhipu2026glm51}, Kimi \citep{moonshot2026kimi26}, Qwen \citep{alibaba2026qwen36}, and MiniMax \citep{minimax2026m27}. Sample is the case-level micro-average. Under Task, the five named columns report task-type pass rates, and Avg.\ is their macro-average. Subtask is the 31-subtask macro-average. The table averages these quantities over three independent runs, and pass@3 reports the fraction of cases solved at least once. This repeated-run metric distinguishes average performance from occasional success.

\paragraph{Overall results}
Claude Opus 4.7 is the strongest model across aggregate views, reaching 65.8\% sample-average success and leaving about one third of the benchmark unsolved. MiMo V2.5 Pro ranks third by sample average but stays competitive under macro-averaging, with a 63.0\% subtask average. GLM 5.1 and Kimi K2.6 are stronger on macro-averaged views than on the raw sample average, showing why long-tail reporting matters for imbalanced releases. The gap between Sample and pass@3 further shows that several agents can solve additional cases across attempts but not reliably.
\begin{figure}[H]
\centering
\includegraphics[width=0.98\columnwidth]{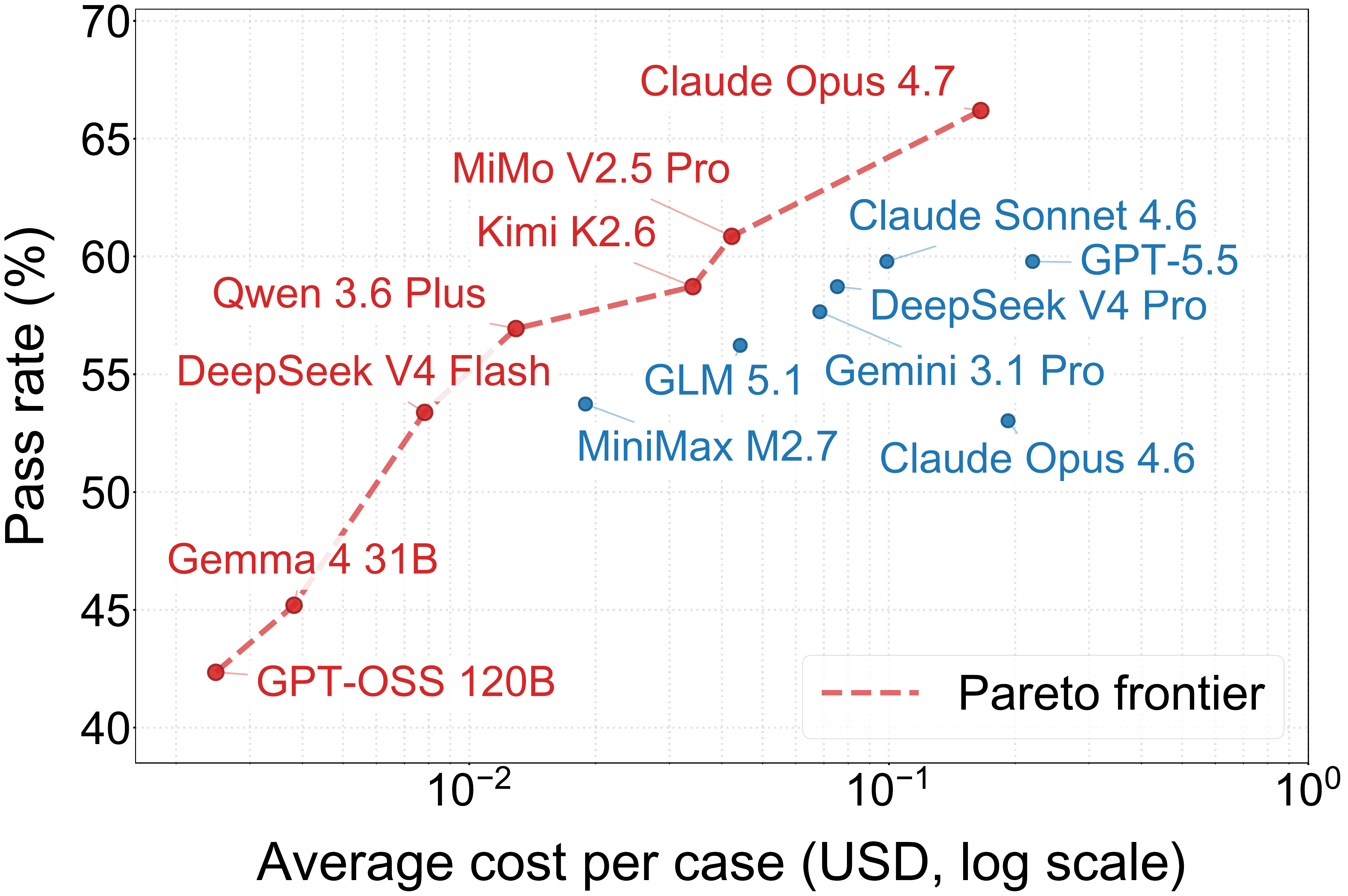}
\caption{Accuracy-cost tradeoff between sample-average pass rate and per-case cost. Higher spending does not directly translate into higher accuracy, and several mid-cost models sit on the frontier.}
\label{fig:exp2-efficiency-cost}
\end{figure}
\paragraph{Efficiency and cost} Figure~\ref{fig:exp2-efficiency-cost} shows that performance and cost are not tightly aligned. Each point is one model. The x-axis is average cost per case on a log scale, the y-axis is sample-average pass rate, and the dashed red line marks the cost-performance frontier. Frontier models are labeled in red, while the rest are labeled in blue. Claude Opus 4.7 costs \$46.54 for a full benchmark run, while GPT-5.5 costs \$61.87 but ranks lower. MiMo V2.5 Pro offers a favorable top-tier tradeoff at \$11.87. The two cheapest models, Gemma 4 31B and GPT-OSS 120B, cost \$1.35 and \$2.04 but trail the top models by more than 20 sample-average points. Appendix~\ref{app:task-subtask-distribution} gives the full subtask robustness matrix and shows where the long-tail ordering differs from the headline leaderboard.

\paragraph{Timeout sensitivity} Re-scoring under alternative caps gives micro-averaged pass rates of $53.3\%$ at 300\,s, $55.8\%$ at 600\,s, and $56.2\%$ at 900\,s over the 11{,}774 runs with valid elapsed times. Tightening to 300\,s censors 301 successes, whereas $50\%$ more budget adds only 38. At 600\,s, 287 runs ($2.4\%$) are timeout-affected, 240 hard stops; because these are right-censored, crediting every one would raise the 900\,s rate by at most $2.4$ points. The top-four ordering is unchanged, so we treat 600\,s as a standardized budget, not a per-task guarantee.

\subsection{Live Benchmarking on Later Logs}

We finally test live benchmarking by rerunning the construction pipeline on two non-overlapping temporal windows: a base window $W_{\mathrm{base}}$ and a later live window $W_{\mathrm{live}}$.
No task labels, quality predicates, rewriting rules, or scorer-construction procedures are changed.
This design tests whether the documented construction procedure remains reusable, rather than whether the same user requests recur.
The filtering, scorer-constructability, and final-selection stages are then applied to $W_{\mathrm{live}}$.
\begin{table}[H]
\centering
\small
\begin{tabular}{lcc}
\toprule
Stage & $W_{\mathrm{base}}$ & $W_{\mathrm{live}}$ \\
\midrule
$N_0^T$ Tool-use & 76,155 & 11,638 \\
$N_1$ Cleaned & 40,713 & 3,505 \\
$N_2$ HQ pool & 6,995 & 606 \\
$N_3$ Scorable & 5,260 & 512 \\
Final & 281 & 34 \\
\bottomrule
\end{tabular}
\caption{Pipeline counts for $W_{\text{base}}$ and $W_{\text{live}}$ under the same protocol. The live window still yields scorer-constructable and final cases, so the pipeline remains usable on later logs without redesign. Final denotes cases remaining after the same final-selection stage.}
\label{tab:exp3-funnel}
\end{table}

\paragraph{Live yield.}
Table~\ref{tab:exp3-funnel} shows that the pipeline remains productive on the later window.
$W_{\mathrm{live}}$ yields 11,638 tool-use sessions, 3,505 cleaned cases, 606 high-quality cases, 512 scorer-constructable cases, and 34 final cases.
The scorer-constructable retention ratio over $N_1$ is 14.6\% in $W_{\mathrm{live}}$, comparable to 12.9\% in $W_{\mathrm{base}}$, supporting the live-yield claim.

\paragraph{Distribution drift.}Table~\ref{tab:exp3-jsd} measures drift along the same four axes as Experiment 1, with all Jensen-Shannon divergence values reported in bits.$B_i$ denotes stage $N_i$ in $W_{\mathrm{base}}$, $L_i$ denotes stage $N_i$ in $W_{\mathrm{live}}$, and $F$ denotes the final release. The live window remains distributionally close to the base window: the largest $W_{\mathrm{live}}$-vs.-$W_{\mathrm{base}}$ divergence is 0.0217 at the cleaned stage and 0.0411 at the scorer-constructable stage. The current final set is also close to both scorer-constructable pools, with all reported JSD values below 0.061 bits. Appendix~\ref{app:live-diagnostics} visualizes the normalized retention and distribution-drift patterns. Thus the later stream is comparable but not identical, motivating versioned live releases.\textsc{RealClawBench} can keep fixed releases for controlled leaderboards and use later versions to track deployed demand.
\begin{table}[t]
\centering
\small
\begin{tabular}{lcccc}
\toprule
Axis & $L_1/B_1$ & $L_3/B_3$ & $F/B_3$ & $F/L_3$ \\
\midrule
Task & 0.0135 & 0.0210 & 0.0186 & 0.0370 \\
Turns & 0.0099 & 0.0195 & 0.0485 & 0.0609 \\
Tools & 0.0195 & 0.0411 & 0.0410 & 0.0315 \\
Calls & 0.0217 & 0.0244 & 0.0227 & 0.0095 \\
\bottomrule
\end{tabular}
\caption{Temporal Jensen-Shannon divergence between $W_{\mathrm{base}}$, $W_{\mathrm{live}}$, and final-release distributions. All reported values remain small, indicating that the live window is comparable to the base release while still reflecting temporal demand shift.}
\label{tab:exp3-jsd}
\end{table}

\section{Conclusion}

We presented \textsc{RealClawBench}, a live-capable framework that projects real OpenClaw sessions into privacy-screened instructions, reconstructed workspaces, and programmatic scorers rather than authored tasks. Aggressive filtering preserves the observable shape of the real-user distribution, the strongest of 14 models reaches only 65.8\% sample-average success, and the same protocol yields 34 new cases from a later log window. Deployment demand can supply a benchmark's source distribution when reconstruction, deterministic scoring, and versioned releases make a changing request stream reproducible.

\section*{Limitations}


This manuscript reports construction fidelity, the main leaderboard, live evaluation, and scorer validation in the appendix. Several limitations remain. First, the benchmark reflects OpenClaw's developer-oriented user population, product surface, tool interface, and logging policy, so it is not a universal estimate of all agent workloads. Second, filtering and reconstruction can still shift the distribution away from raw usage. Tasks depending on private services, unreleasable files, credentials, GUI state, or unstable external resources are more likely to be removed or simplified. Third, programmatic scorers favor outcomes that can be checked in a final workspace or bounded execution trace. They may miss code maintainability, design quality, security posture, and partial but useful progress. Fourth, model scores are conditioned on one harness, timeout, tool interface, and cost model. Different budgets or tool implementations could change absolute scores even with the construction protocol fixed. Finally, live releases require governance so that privacy review, versioning, scorer audits, and leaderboard comparability remain reliable as the data stream changes.

\section*{Ethical Considerations}



\paragraph{Data source and authorization.}
All source sessions came from employees of the organization operating the internal OpenClaw deployment, during ordinary work. No external-customer, third-party, or public-user data were included. Collection and research use were authorized in advance under the organization's employee-data governance, privacy, and compliance procedures, and the release was approved through internal privacy, compliance, and ethics review. These procedures are referenced by items D3 and D4 of the Responsible NLP checklist.

\paragraph{Screening and sanitization.}
Raw logs remained access-controlled and unreleased; only sanitized derivatives entered the benchmark. Stage~1 screened every candidate for personal identifiers, credentials, tokens, private URLs and service endpoints, proprietary or harmful content, and licensing risk. Private names, paths, accounts, services, and organization-specific context were removed or replaced, and workspaces were sanitized, reconstructed from non-sensitive evidence, or substituted with behavior-preserving fixtures. Every package then received human privacy review; cases that could not be made safe without altering their meaning were rejected, not paraphrased.

\paragraph{Post-release audit.}
Because the released artifact, not the upstream detector, carries the disclosure risk, we audited all 281 packages after de-identification, covering every instruction, workspace, fixture, verifier, metadata record, and manifest. The audit found no unresolved personal identifiers, credentials, employee identifiers, proprietary files, or private service traces. Each release card records its privacy-review status, and the same screening, human review, and post-release audit apply to every subsequent versioned release.

\bibliography{custom}
\newpage
\appendix

\section{Experimental Configuration}
Table~\ref{tab:experimental-config} summarizes the main settings used in our
experiments. We include the benchmark size, execution environment, model-serving
setup, timeout limits, and sampling policy. To preserve
anonymity, the table omits local file paths, private service
identifiers, and other deployment-specific details that are not needed for
reproducing the reported experimental protocol.
These settings should be read as fixed control variables for the leaderboard: each model receives the same workspace reset, tool interface, timeout, and sampling configuration.
The hardware rows describe the shared evaluation infrastructure rather than model-specific resources.
The table is therefore intended to make the run conditions auditable, not to define a new experimental variable.
Cost estimates in the main text are computed from recorded input and output token counts and the model-specific prices used at evaluation time.
For locally served open-weight models, the reported costs should be interpreted as evaluation-accounting estimates rather than provider billing receipts.

\begin{table}[H]
\centering
\small
\setlength{\tabcolsep}{5pt}
\begin{tabular}{l p{0.58\linewidth}}
\toprule
Item & Configuration \\
\midrule
Benchmark size & 281 tasks, 14 models \\
Runtime & OpenClaw agent runtime \\
Workspace & One isolated workspace per task \\
Scoring & Case-specific Python verifiers \\
Allowed tools &
\begin{tabular}[t]{@{}l@{}}
File: read, write, edit \\
Search: file search, file list \\
Execution: shell command
\end{tabular} \\
CPU machines & 20 machines, 32 vCPUs each \\
Controller CPU & 2 $\times$ AMD EPYC 7742, 256 logical cores \\
Controller memory & 2.0 TiB RAM \\
GPU server & 8 $\times$ NVIDIA A100-SXM4-80GB \\
Operating system & Ubuntu 24.04 \\
Python version & Python 3.12 \\
Node.js version & Node.js 22.x in Docker \\
Local model server & vLLM 0.19.1 \\
Task timeout & 600 s\\
Sampling &
\begin{tabular}[t]{@{}l@{}}
temperature = 1.0 \\
top\_p = 1.0 \\
top\_k = -1 \\
min\_p = 0.0
\end{tabular} \\
\bottomrule
\end{tabular}
\caption{Experimental configuration for the main evaluation.}
\label{tab:experimental-config}
\end{table}

\begin{table*}[t]
\centering
\small
\begin{tabular}{llp{0.64\textwidth}}
\toprule
Stage & Description & Main Reason \\
\midrule
$N_0^A$ & Raw API calls & Source OpenClaw API-call logs. \\
$N_0^S$ & Unique user sessions & Fold by 5-turn-prefix hash. Remove sessions whose user turns are all framework boilerplate. \\
$N_0^T$ & Tool-using sessions & Keep sessions with at least one defined tool. This is the primary real-user reference for agent benchmarking. \\
$N_1$ & Framework-cleaned & Remove framework injection, system templates, memory jobs, cron messages, subagent dispatches, and related non-user turns. \\
$N_2$ & High-quality pool & Enforce request length, intent signal, no secrets, within-label deduplication, and removal of continuation, meta, and greeting-only sessions. \\
$N_3$ & Scorer-constructable & Drop failed or unclear sessions without a verifiable final state. \\
$C$ & Candidate case cards & Construct task, rubric, seed workspace, and verifier. Exclude aggregate markdown files. \\
$F$ & Final evaluation cases & Manual review removes ambiguous, unstable, unreproducible, or low-quality cases. \\
\bottomrule
\end{tabular}
\caption{Construction-funnel details with stage descriptions and main filtering reasons. The table clarifies what each stage removes and why the large reductions correspond to reproducibility, privacy, quality, and scorer-constructability requirements.}
\label{tab:appendix-construction-funnel}
\end{table*}

\section{Construction Funnel Details}
\label{app:construction-funnel}

Table~\ref{tab:appendix-construction-funnel} expands the compact funnel in Table~\ref{tab:exp1-funnel} by recording each stage's role and main filtering reason.
$N_0^A$ and $N_0^S$ are log-accounting stages, while $N_0^T$ defines the tool-using reference population used in the construction-fidelity experiment.
The later stages are benchmark-construction gates: $N_1$ removes framework artifacts, $N_2$ keeps sessions with sufficient intent signal, $N_3$ requires plausible automatic verification, and $C$ and $F$ turn records into reviewed benchmark artifacts.
The large drops therefore reflect the cost of turning private, stateful production sessions into public, reproducible, and scorable tasks. The key point is that most filtering is tied to reproducibility and verification requirements rather than to late manual rebalancing. Future releases can report the same funnel so observed changes can be attributed to user-demand drift or explicit construction-policy changes.

\section{Task and Subtask Distribution}
\label{app:task-subtask-distribution}
\begin{figure*}[t]
\centering
\includegraphics[width=0.98\textwidth]{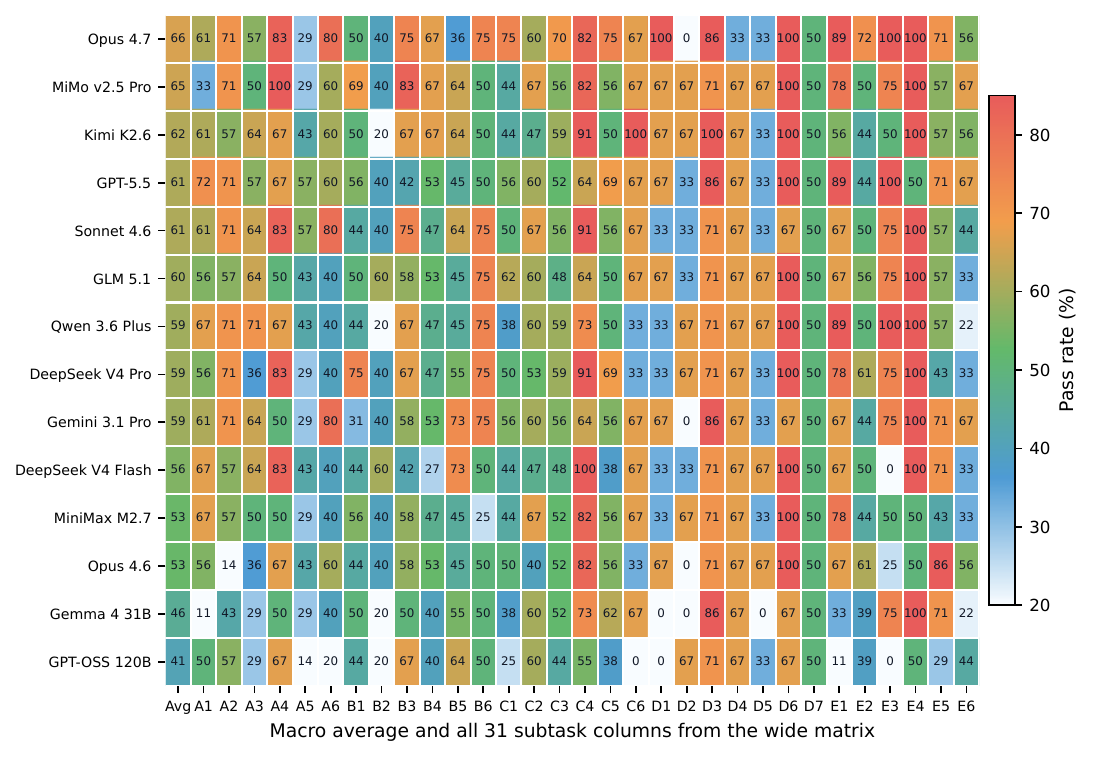}
\caption{Subtask robustness matrix showing model pass rates across the full subtask taxonomy. The matrix reveals that rankings can shift under long-tail macro-averaging, so headline sample accuracy alone is not sufficient.}
\label{fig:exp2-subtask-matrix}
\end{figure*}

Table~\ref{tab:appendix-task-distribution} reports the complete task and subtask composition of the final evaluation set.
Share of all is computed over the full final set, and share of task is computed within the corresponding top-level task type.
The task-type rows correspond to the five inner-ring categories in Figure~\ref{fig:exp1-task-distribution}, while the subtask rows correspond to the outer-ring slices.
The taxonomy is operational rather than purely semantic: top-level task types describe what the agent must do in the environment, and subtasks provide finer labels for sampling, macro-averaging, and error analysis.
These labels are metadata only and are not shown to the evaluated agent.
The distribution is visibly imbalanced, which is expected for a benchmark distilled from real usage.
The small subtasks should therefore be read as diagnostic slices rather than independently powered benchmarks.
\begin{table*}[t]
\centering
\small
\setlength{\tabcolsep}{4pt}
\begin{tabularx}{\textwidth}{p{0.33\textwidth}Xrrr}
\toprule
Task/Subtask & Description & Cases & Share of all & Share of task \\
\midrule
\texttt{A.file\_create} & File creation & 57 & 20.3\% & -- \\
\quad\texttt{A1.single\_html\_game} & Single-file HTML game & 18 & 6.4\% & 31.6\% \\
\quad\texttt{A3.code\_module} & Code module or script & 14 & 5.0\% & 24.6\% \\
\quad\texttt{A2.web\_app\_page} & Web app page or courseware & 7 & 2.5\% & 12.3\% \\
\quad\texttt{A5.doc\_report} & Document or report & 7 & 2.5\% & 12.3\% \\
\quad\texttt{A4.manim\_video\_script} & Animation or video script & 6 & 2.1\% & 10.5\% \\
\quad\texttt{A6.binary\_data\_file} & Data or office file & 5 & 1.8\% & 8.8\% \\
\addlinespace
\texttt{B.code\_fix} & Code fixing & 63 & 22.4\% & -- \\
\quad\texttt{B1.bug\_fix} & Runtime or compile bug fix & 16 & 5.7\% & 25.4\% \\
\quad\texttt{B4.refactor} & Refactoring and code quality & 15 & 5.3\% & 23.8\% \\
\quad\texttt{B3.feature\_tweak} & Small feature adjustment & 12 & 4.3\% & 19.0\% \\
\quad\texttt{B5.feature\_add} & New feature or module & 11 & 3.9\% & 17.5\% \\
\quad\texttt{B2.config\_fix} & Configuration fix & 5 & 1.8\% & 7.9\% \\
\quad\texttt{B6.review\_report} & Code or document review & 4 & 1.4\% & 6.3\% \\
\addlinespace
\texttt{C.data\_query} & Data and codebase querying & 88 & 31.3\% & -- \\
\quad\texttt{C3.behavior\_logic\_analysis} & Behavior or logic analysis & 27 & 9.6\% & 30.7\% \\
\quad\texttt{C1.locate\_specific\_code} & Locate specific code or references & 16 & 5.7\% & 18.2\% \\
\quad\texttt{C5.extract\_structured\_data} & Extract structured data & 16 & 5.7\% & 18.2\% \\
\quad\texttt{C2.codebase\_inventory\_report} & Codebase inventory report & 15 & 5.3\% & 17.0\% \\
\quad\texttt{C4.print\_file\_contents} & Print file contents & 11 & 3.9\% & 12.5\% \\
\quad\texttt{C6.system\_env\_inspection} & System or environment inspection & 3 & 1.1\% & 3.4\% \\
\addlinespace
\texttt{D.cmd\_exec} & Command execution & 24 & 8.5\% & -- \\
\quad\texttt{D3.write\_script\_automation} & Write automation script & 7 & 2.5\% & 29.2\% \\
\quad\texttt{D1.git\_clone\_inspect} & Git clone and inspect changes & 3 & 1.1\% & 12.5\% \\
\quad\texttt{D2.fs\_inventory\_analyze} & Filesystem inventory and analysis & 3 & 1.1\% & 12.5\% \\
\quad\texttt{D4.create\_files\_config} & Create files or edit config & 3 & 1.1\% & 12.5\% \\
\quad\texttt{D5.install\_setup\_env} & Package installation or environment setup & 3 & 1.1\% & 12.5\% \\
\quad\texttt{D6.diagnose\_cleanup} & Diagnosis and cleanup & 3 & 1.1\% & 12.5\% \\
\quad\texttt{D7.doc\_convert\_audit} & Document conversion or audit & 2 & 0.7\% & 8.3\% \\
\addlinespace
\texttt{E.project\_build} & Project building & 49 & 17.4\% & -- \\
\quad\texttt{E2.module\_feature\_impl} & Module or feature implementation & 18 & 6.4\% & 36.7\% \\
\quad\texttt{E1.web\_app\_prototype} & Web-app prototype or game & 9 & 3.2\% & 18.4\% \\
\quad\texttt{E6.spec\_doc\_update} & Specification or documentation update & 9 & 3.2\% & 18.4\% \\
\quad\texttt{E5.audit\_review\_report} & Code or policy audit report & 7 & 2.5\% & 14.3\% \\
\quad\texttt{E3.skill\_plugin\_pack} & Skill or plugin packaging & 4 & 1.4\% & 8.2\% \\
\quad\texttt{E4.scaffold\_setup} & Project scaffolding setup & 2 & 0.7\% & 4.1\% \\
\midrule
\texttt{TOTAL} & Total & 281 & 100.0\% & -- \\
\bottomrule
\end{tabularx}
\caption{Complete task and subtask distribution for the final evaluation set, including within-task and overall shares. The table makes the real-usage imbalance explicit and explains the task and subtask labels used for macro-averaged reporting.}
\label{tab:appendix-task-distribution}
\end{table*}
\texttt{C.data\_query} is the largest group, accounting for 31.3\% of the release, while \texttt{D.cmd\_exec} is the smallest top-level group at 8.5\%. Several subtasks contain only two to five cases, which is why the main results report both sample-average pass rate and macro-averages over task types and subtasks. The task-type columns in the leaderboard correspond to the five top-level task rows, while the subtask matrix expands them into the 31 outer-ring labels. The three largest top-level categories, data/codebase querying, code fixing, and file creation, account for 74.0\% of the final set, so a pure sample average would be strongly shaped by these workflows. At the subtask level, the largest slice is behavior or logic analysis with 27 cases, while multiple command-execution and project-building subtasks contain only two or three cases.
Because the benchmark is not manually balanced, the sample average estimates performance under the released distribution, the five-way task average reduces dominance by the largest top-level category, and the 31-way subtask average highlights long-tail robustness.
Figure~\ref{fig:exp2-subtask-matrix} reports the full model-by-subtask matrix used for this analysis.
Its first column is the macro-average over all subtasks, while the remaining columns show pass rates for each subtask separately.
This view is larger than the main leaderboard, so we place it in the appendix, but it is important for interpreting close model comparisons.
The ordering is close to the main table but not identical: MiMo V2.5 Pro moves closer to Claude Opus 4.7 under subtask macro-averaging, and several mid-table comparisons shift when rare subtasks receive equal weight.
The matrix also makes clear that high overall performance can hide uneven behavior across task families.
For example, a model can perform well on frequent data-querying cases while losing ground on smaller creation, configuration, or command-execution subtasks.
This supports reporting sample, task, and subtask averages together rather than treating the micro-average as the only benchmark summary.

\section{Live Benchmark Diagnostics}
\label{app:live-diagnostics}

\begin{figure}[H]
\centering
\includegraphics[width=0.98\columnwidth]{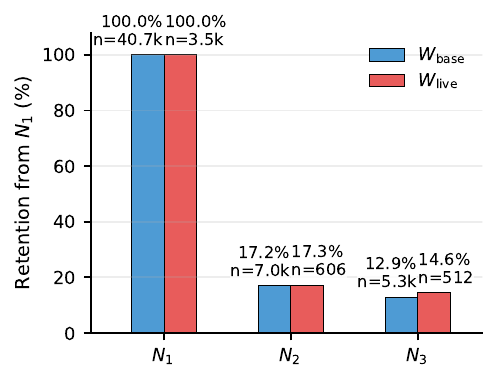}
\caption{Live-window retention after normalizing each temporal window by its $N_1$ cleaned count. The live window retains a comparable scorer-constructable share to the base window, supporting the live-yield claim.}
\label{fig:exp3-live-summary}
\end{figure}

\begin{figure*}[t]
\centering
\includegraphics[width=0.98\textwidth]{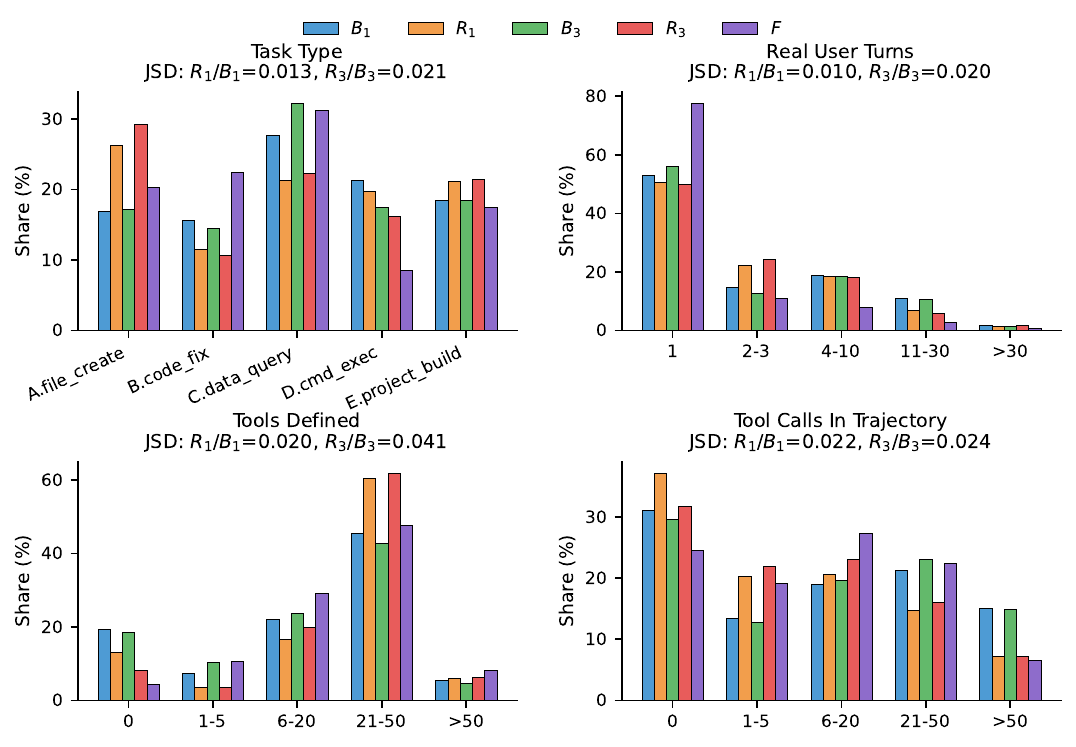}
\caption{Distribution drift under live benchmarking. The figure tests whether the later window remains comparable to the base window across task type, user turns, tools defined, and tool calls. The distributions remain close across all four axes, showing that the live stream preserves the benchmark's source distribution while still capturing temporal variation.}
\label{fig:exp3-distribution-drift}
\end{figure*}

The main text reports the live-window counts and JSD values in compact tables.
Here we include the corresponding visual diagnostics for the same temporal split.
Figure~\ref{fig:exp3-live-summary} normalizes each window by its $N_1$ cleaned count and reports the retained share at the $N_2$ high-quality and $N_3$ scorer-constructable stages.
This view makes the yield comparison easier to read: although $W_{\mathrm{live}}$ is smaller in absolute size, its scorer-constructable retention remains comparable to $W_{\mathrm{base}}$. Figure~\ref{fig:exp3-distribution-drift} complements Table~\ref{tab:exp3-jsd} by showing the same drift comparisons across task type, user turns, tools defined, and tool calls. In normalized terms, $W_{\mathrm{live}}$ keeps 17.3\% of cleaned cases at $N_2$ and 14.6\% at $N_3$, close to 17.2\% and 12.9\% in $W_{\mathrm{base}}$. Thus the later window is smaller mainly because its log span is shorter, not because the construction pipeline stops finding high-quality or scorable sessions. The bars compare the base and live windows at $N_1$ and $N_3$ together with the current final release.
Across all four axes, the distributions remain close enough for cross-release comparison while still showing measurable temporal variation. The largest base-live drift is 0.0411 bits for tools defined at $N_3$, and the final set remains within 0.061 bits of the live-window scorer-constructable pool on all reported axes.

\section{Extra-Quality Live Subset}
\label{app:extra-quality-live-subset}

The 34 final cases from $W_{\mathrm{live}}$ provide a compact extra-quality subset for checking whether the same evaluation protocol continues to separate models on newly constructed tasks.
Table~\ref{tab:extra34-live-leaderboard} reports the corresponding leaderboard.
\begin{table}[H]
\centering
\small
\begin{tabular}{l l c c c}
\toprule
Rank & Model & Sample & Task & pass@3 \\
\midrule
1 & GPT-5.5 & \first{71.6\%} & \first{75.3\%} & \first{73.5\%} \\
2 & Claude Opus 4.7 & {61.8\%} & {67.9\%} & 70.6\% \\
3 & Kimi K2.6 & {61.8\%} & 61.2\% & {73.5\%} \\
4 & Claude Sonnet 4.6 & 60.8\% & 57.8\% & {73.5\%} \\
5 & Claude Opus 4.6 & 59.8\% & 60.5\% & 70.6\% \\
6 & Gemini 3.1 Pro & 58.8\% & 63.7\% & 70.6\% \\
7 & MiMo V2.5 Pro & 58.8\% & 58.5\% & 76.5\% \\
8 & GLM 5.1 & 57.8\% & 59.9\% & 70.6\% \\
9 & DeepSeek V4 Pro & 56.9\% & 55.4\% & 64.7\% \\
10 & Qwen 3.6 Plus & 54.9\% & 54.5\% & 64.7\% \\
11 & MiniMax M2.7 & 52.9\% & 52.8\% & 61.8\% \\
12 & DeepSeek V4 Flash & 52.0\% & 51.4\% & 61.8\% \\
13 & Gemma 4 31B & 49.0\% & 47.1\% & 55.9\% \\
14 & GPT-OSS 120B & 44.1\% & 45.8\% & 61.8\% \\
\bottomrule
\end{tabular}
\caption{Leaderboard on the 34-case extra-quality live subset. The subset acts as a live sanity check and continues to separate models, with GPT-5.5 leading on both Sample and Task.}
\label{tab:extra34-live-leaderboard}
\end{table}
Sample follows the main table's micro-average definition, Task is the task-type macro average, and pass@3 reports the fraction of cases solved at least once. GPT-5.5 leads this subset on both Sample and Task, while Claude Opus 4.7 and Kimi K2.6 form a close second tier with identical Sample but different macro profiles. MiMo V2.5 Pro has the highest pass@3, suggesting some failures are recoverable across repeated runs. The spread between top and bottom models is 27.5 Sample points, so even this small subset remains discriminative. The gap between Sample and pass@3 is also informative: MiMo V2.5 Pro solves at least one run on 76.5\% of cases but averages 58.8\%, indicating many non-repeatable successes. Because this subset is small, we use it as a live sanity check, not a replacement for the main leaderboard.

\section{Environment and Scorer Validation}
\label{app:scorer-validation}

The main leaderboard uses case-specific rule-based programmatic verifiers rather than LLM judges.
This appendix asks a separate validation question: whether an LLM auditor can be reliable when it is given execution evidence.
The goal is to test the value of execution evidence for LLM auditing, not to replace the rule-based verifiers used for model ranking.
We compare human judgment with the \textsc{RealClawBench} full-evidence LLM auditor and an output-only LLM judge baseline on the final benchmark.
The output-only baseline receives only the task, rubric, and final model output.
\begin{figure}[H]
\centering
\includegraphics[width=0.98\columnwidth]{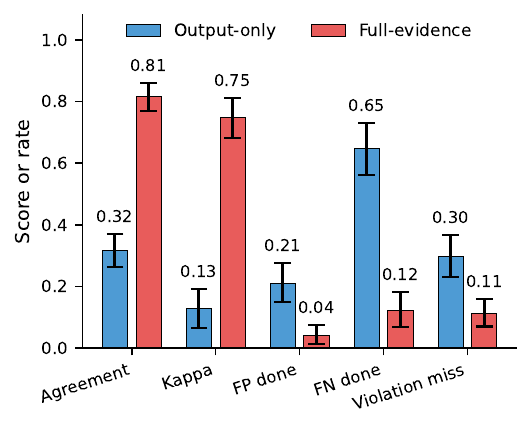}
\caption{Scorer-validation metrics comparing agreement and benchmark-critical error rates. Full-evidence auditing improves agreement with human labels while reducing false-positive done, false-negative done, and violation-miss errors.}
\label{fig:appendix-scorer-metrics}
\end{figure}
\begin{figure*}[t]
\centering
\includegraphics[width=0.95\textwidth]{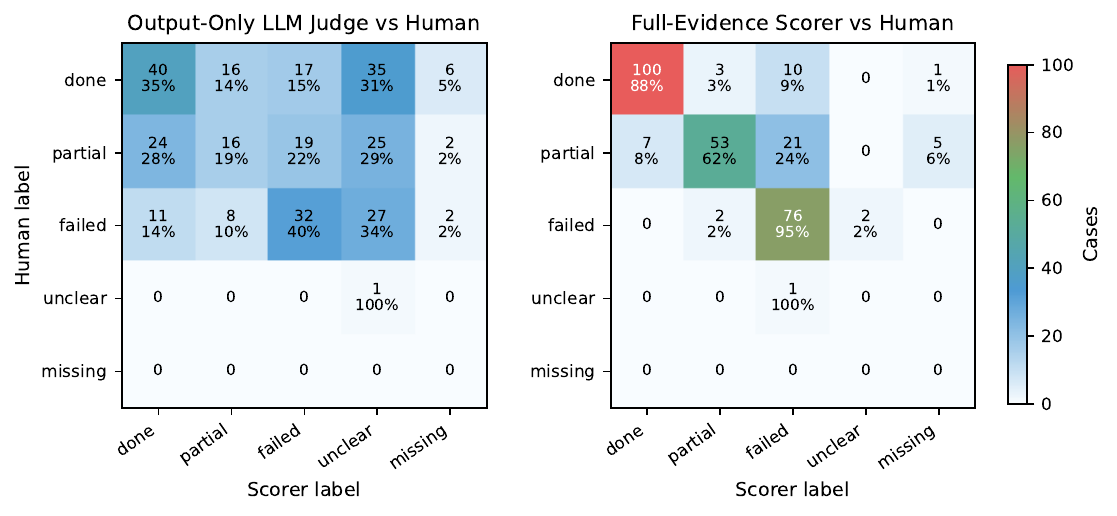}
\caption{Verdict confusion matrices with human labels as rows and scorer labels as columns. The full-evidence auditor is more concentrated on the diagonal, while output-only judging more often confuses done, partial, failed, and unclear outcomes.}
\label{fig:appendix-confusion}
\end{figure*}
The full-evidence LLM auditor additionally sees verifier checks, tool summaries, runtime metadata, and reconstructed task evidence.
This matters because many agent outcomes are invisible from the final answer alone: an agent may claim edits it never made, ignore required evidence, or fail after a command error while still producing a confident response.
The human reference separates successful completion, partial progress, failed attempts, unclear outcomes, and missing outputs.
We report exact-verdict agreement and Cohen's $\kappa$ for overall alignment, plus false-positive done, false-negative done, and violation-miss rates for benchmark-critical error modes.
Lower is better for the three error rates. Figure~\ref{fig:appendix-scorer-metrics} first summarizes the main pattern: the full-evidence LLM auditor improves agreement with human labels while reducing all three failure modes, rather than merely becoming more permissive.

\begin{table}[H]
\centering
\small
\begin{tabular}{lcc}
\toprule
Metric & Output-only & Full-evidence \\
\midrule
Compared cases & 281 & 281 \\
Exact-verdict agreement & 0.317 & \textbf{0.815} \\
Cohen's $\kappa$ & 0.128 & \textbf{0.748} \\
False-positive done rate & 0.210 & \textbf{0.042} \\
False-negative done rate & 0.649 & \textbf{0.123} \\
Violation miss rate & 0.297 & \textbf{0.114} \\
\bottomrule
\end{tabular}
\caption{Scorer-validation against human labels for output-only and full-evidence LLM auditing. Full-evidence auditing achieves higher agreement and lower error rates, showing the value of execution evidence.}
\label{tab:appendix-scorer-validation}
\end{table}
Table~\ref{tab:appendix-scorer-validation} then gives the exact values. The full-evidence LLM auditor reaches 81.5\% exact-verdict agreement and $\kappa=0.748$, compared with 31.7\% and $\kappa=0.128$ for the output-only baseline.
It also reduces false-positive done from 21.0\% to 4.2\% and false-negative done from 64.9\% to 12.3\%, supporting execution-aware LLM auditing for stateful agent tasks. In absolute terms, execution evidence improves exact agreement by 49.8 percentage points and raises Cohen's $\kappa$ by 0.620. The error reductions are benchmark-critical: false-positive done labels fall by a factor of five, and false-negative done labels fall by more than fivefold, which means the auditor is less likely to both over-credit and under-credit completed agent work.
Figure~\ref{fig:appendix-confusion} gives the label-level view. Rows are human labels and columns are auditor labels.
The full-evidence LLM auditor is much more concentrated on the diagonal, while the output-only judge frequently confuses \texttt{done}, \texttt{partial}, \texttt{failed}, and \texttt{unclear} cases.

\section{Case Study: Recovery after a Failed Edit}
We use one representative code-repair case to show how execution traces expose reliability differences that final pass rates alone hide.
The task is to fix a sensitive-information leak in \texttt{tools/lark\_auth.py}.
In \texttt{device\_flow\_authorize()}, the original code interpolates the raw API message from \texttt{resp.get("msg")} into an exception when Device Flow initialization fails.
Because this message may contain user access or refresh tokens, the correct repair must route it through the existing \texttt{\_sanitize\_lark\_error()} helper before raising the exception.
The verifier requires the target file to exist, core symbols such as \texttt{device\_flow\_authorize}, \texttt{NeedsAuthError}, \texttt{get\_user\_token}, and \texttt{\_sanitize\_lark\_error} to be preserved, the raw \texttt{\{resp.get("msg")\}} interpolation to disappear, and the Device Flow initialization failure path to still raise an error.
Thus, the case requires a small but precise code change, not a high-level explanation, while preserving surrounding authentication logic.

Across repeated runs, 18 trajectories passed and 26 failed.
Passing traces were not simply traces without tool errors.
One GPT-5.5 trajectory first failed an exact edit because the supplied text did not match the file contents, receiving a \texttt{old text not found} error.
The agent then re-read the file, located the actual surrounding code, rewrote the exception construction, ran the tests, and printed the required confirmation after all seven tests passed.
This trajectory illustrates a successful recovery pattern: the first edit failed, but the agent used tool feedback to revise its plan and complete the task.

The failed trajectories show several distinct weaknesses.
Some models made the sanitization change but omitted the required confirmation string, causing verifier failure even though the edit was mostly correct.
Others changed exception text or control flow in a way that broke the expected Device Flow failure path.
More severe failures occurred when agents tried to repair a truncated or partially read file and lost core functions such as \texttt{device\_flow\_authorize}.
These runs often ended with plausible summaries claiming that the leak was fixed, while the resulting file no longer contained all required authentication symbols.
Local-model runs also showed path-grounding and continuation failures, including reading from the wrong workspace path or stopping after partial tool actions.
Table~\ref{tab:lark-auth-case-study} summarizes these recovery and failure patterns.
\begin{table}[t]
\centering
\small
\setlength{\tabcolsep}{5pt}
\begin{tabular}{p{0.28\linewidth} p{0.62\linewidth}}
\toprule
Aspect & Observation \\
\midrule
Task & Sanitize a raw Lark API error message before raising a Device Flow error \\
Target file & \texttt{tools/lark\_auth.py} \\
Required behavior & Use \texttt{\_sanitize\_lark\_error()} while preserving authentication logic \\
Verification & Static symbol checks, leak-pattern check, preserved raise path, test execution, and required stdout confirmation \\
Successful recovery & Failed exact edit, re-read file, applied corrected patch, ran tests, printed confirmation \\
Common failures & Missing confirmation, changed error path, lost core functions, wrong path, incomplete continuation \\
Capability tested & Tool recovery, precise code editing, state preservation, and verification discipline \\
\bottomrule
\end{tabular}
\caption{Recovery and failure patterns in the Lark authentication repair case. The case shows that success depends not only on knowing the required patch, but also on recovering from tool errors, preserving nearby code, and completing verification.}
\label{tab:lark-auth-case-study}
\end{table}
Overall, the case shows that failure is not always caused by missing domain knowledge.
Many models understood that the raw API message should be sanitized, but success depended on execution reliability: recovering from edit mismatches, preserving nearby code, running tests, and satisfying the explicit completion contract.
This is a realistic agentic failure mode in existing projects, where a small patch requires robust tool use, local state tracking, and post-edit verification.

\section{Realness Metric Computation}
\label{app:realness-metrics}
Algorithm~\ref{alg:realness-metrics} specifies how the metrics in Table~\ref{tab:realism-gap} are computed. It maps both the deployed reference stream and each benchmark to a shared taxonomy before computing JSD and TV\@. When comparable public prompts are available, it applies rule-based text-signal functions to compute the intent score and selected signal shares.
\begin{algorithm*}[t]
\caption{Computing Realness Metrics}
\label{alg:realness-metrics}
\begin{algorithmic}[1]
\Require Deployed OpenClaw tool-use sessions $R$, benchmark artifacts $\{B_i\}$, shared taxonomy $\mathcal{C}$, and text-signal functions $\mathcal{F}=\{f_1,\ldots,f_8\}$
\Ensure JSD, TV, Intent score, and selected signal shares for each benchmark $B_i$
\State Map each real session in $R$ to a category $c\in\mathcal{C}$
\State Count $n_R(c)$ and normalize $p_R(c)\gets n_R(c)/\sum_{c'\in\mathcal{C}}n_R(c')$
\ForAll{benchmarks $B_i$}
    \State Map each task in $B_i$ to a category $c\in\mathcal{C}$
    \State Count $n_{B_i}(c)$ and normalize $p_{B_i}(c)\gets n_{B_i}(c)/\sum_{c'\in\mathcal{C}}n_{B_i}(c')$
    \State $m \gets (p_R+p_{B_i})/2$
    \State $\mathrm{JSD}_i \gets \frac{1}{2}\mathrm{KL}(p_R\|m)+\frac{1}{2}\mathrm{KL}(p_{B_i}\|m)$
    \State $\mathrm{TV}_i \gets \frac{1}{2}\sum_{c\in\mathcal{C}} |p_R(c)-p_{B_i}(c)|$
    \If{$B_i$ exposes comparable public task prompts}
        \ForAll{task prompts $x\in B_i$}
            \For{$j=1$ to $8$}
                \State Compute binary signal $f_j(x)\in\{0,1\}$
            \EndFor
        \EndFor
        \State $\mathrm{Intent}_i \gets |B_i|^{-1}\sum_{x\in B_i}\sum_{j=1}^{8} f_j(x)$
        \State $\mathrm{Share}_{i,j} \gets |B_i|^{-1}\sum_{x\in B_i} f_j(x)$ for selected signals $j$
    \Else
        \State Mark text-realism metrics as unavailable
    \EndIf
\EndFor
\end{algorithmic}
\end{algorithm*}
\section{Representative Task Cards}
\label{app:representative-task-cards}

This appendix provides representative task cards. Each card summarizes the workspace setup, agent objective, expected artifacts, constraints, and oracle-grading signal for one task family.
Table~\ref{tab:appendix-representative-manim-task-card} shows a Manim educational video script workflow from the file-creation task type.
The example illustrates how a raw session is converted into a standalone instruction, a bounded input workspace, and concrete verifier checks.
It also shows that the released task exposes the necessary evidence without revealing the original private session. The verifier focuses on observable artifacts, such as file placement, syntax validity, required scene classes, configuration, and content signals, rather than subjective judgments of animation quality.
\begin{table*}[t]
\centering
\small
\begin{tabularx}{\textwidth}{p{0.18\textwidth}X}
\toprule
\textbf{Field} & \textbf{Description} \\
\midrule
Task ID &
\texttt{04212e3aa859be28} \\

Subclass &
\texttt{A.file\_create}. Tier \texttt{tier2\_pastes\_inline}. Task tags \texttt{[file\_or\_workspace, artifact]} \\

Task &
\begin{minipage}[t]{\linewidth}
\textbf{Chain Rule Manim educational video script.}
Create a Manim Community Edition v0.19.1 script for an educational video explaining the chain rule.
The workspace already provides the storyboard file \texttt{root/backprop-demo/chain\_rule/storyboard.py}. Refer to its scene descriptions, characters, and configuration information, but do not need to copy it exactly.

\par\medskip
\textbf{Output file path.}
The script must be saved to the following path relative to the workspace root:
\texttt{gradient-visualization/gradient\_chain/scenes.py}.
Any other location is considered non-compliant.

\par\medskip
\textbf{Required Scene classes.}
Based on storyboard scenes \texttt{S01}, \texttt{S04}, \texttt{S05}, \texttt{S06}, \texttt{S07}, \texttt{S08}, and \texttt{S10}, create the corresponding Manim \texttt{Scene} classes.
Each class must directly inherit from \texttt{Scene}.
Class names must use the format \texttt{S\{ID\}\_\{short\_desc\}}.

\begin{enumerate}[leftmargin=*, nosep]
    \item \texttt{S01\_SimpleNetwork}: corresponds to storyboard scene S01. Show a simple multi-layer neural network, with data flowing from left to right and a large X displayed at the output.
    \item \texttt{S04\_DominoToNetwork}: corresponds to S04. Domino tiles transform into neural-network layers, with at least one transformation animation.
    \item \texttt{S05\_ChainRuleFormula}: corresponds to S05. Animate the chain-rule formula \texttt{dy/dx = dy/du · du/dx}, rendered using \texttt{MathTex} or \texttt{Tex}.
    \item \texttt{S06\_FormulaWithNetwork}: corresponds to S06. Overlay the chain-rule formula on the neural-network diagram so that the formula and the network are visible at the same time.
    \item \texttt{S07\_ConcreteExample}: corresponds to S07. Show a concrete composite-function derivative example, such as the differentiation process for \texttt{(3x+1)\textasciicircum 4}.
    \item \texttt{S08\_ComputeGradient}: corresponds to S08. Use the chain rule to compute the gradient of a certain layer and reflect the idea of error backpropagation.
    \item \texttt{S10\_GradientDescentStep}: corresponds to S10. Demonstrate one gradient-descent update step and show how parameters are updated according to the chain-rule gradient.
\end{enumerate}

\par\medskip
If a storyboard scene does not contain enough content to form a complete animation, the agent may extend it appropriately, but the class names and scene themes must still match.

\par\medskip
\textbf{Global configuration.}
In the global scope of the script, set Manim configuration to match \texttt{VIDEO\_CONFIG} in \texttt{storyboard.py}:
\texttt{config.pixel\_width = 1280}, \texttt{config.pixel\_height = 720}, and \texttt{config.background\_color = WHITE} or \texttt{"\#FFFFFF"}.

\par\medskip
\textbf{Content requirements.}
The script must contain Chinese characters, such as character dialogue or explanatory text.
It must use \texttt{MathTex} or \texttt{Tex} at least once to render a mathematical formula.
It is recommended, but not required, to use colors such as \texttt{\#E53935}, \texttt{\#1E88E5}, and \texttt{\#43A047} to highlight key elements.
\end{minipage}
\\

Input workspace &
\begin{minipage}[t]{\linewidth}
Files placed in the workspace before the agent runs:
\begin{itemize}[leftmargin=*, nosep]
    \item \texttt{gradient-visualization/README.md} (231 B)
    \item \texttt{gradient-visualization/gradient\_chain/init.py} (72 B)
    \item \texttt{root/backprop-demo/chain\_rule/README.md} (267 B)
    \item \texttt{root/backprop-demo/chain\_rule/storyboard.py} (3807 B)
\end{itemize}
\end{minipage}
\\

Rubric &
\begin{minipage}[t]{\linewidth}
Must pass:
\begin{enumerate}[leftmargin=*, nosep]
    \item \texttt{file\_exists} --- \texttt{gradient-visualization/gradient\_chain/scenes.py} exists and is nonempty.
    \item \texttt{syntax\_valid} --- the file can be compiled with \texttt{py\_compile}.
    \item \texttt{manim\_import} --- the file contains \texttt{from manim import *} or an equivalent Manim import.
    \item \texttt{scene\_classes\_defined} --- at least six of the seven required \texttt{Scene} classes are defined, and these classes inherit from \texttt{Scene}.
    \item \texttt{config\_set} --- \texttt{pixel\_width=1280} and \texttt{pixel\_height=720} are set, and a \texttt{WHITE} background is used.
    \item \texttt{content\_richness} --- the file contains Chinese characters and uses \texttt{MathTex} or \texttt{Tex} at least once.
    \item \texttt{has\_main\_block} --- the bottom of the file contains an \texttt{if \_\_name\_\_ == "\_\_main\_\_":} block.
\end{enumerate}
\end{minipage}
\\
\bottomrule
\end{tabularx}
\caption{Representative task card showing the released instruction, workspace evidence, and verifier checks. The example illustrates how a private session is projected into a standalone, bounded, and automatically graded benchmark instance.}
\label{tab:appendix-representative-manim-task-card}
\end{table*}

\section{Use of AI}

We used large language models (LLMs) throughout the research and writing process as auxiliary tools for language polishing, grammar correction, and clarity improvement. In particular, LLM-based assistants were used to refine sentence structure, improve academic writing style, and identify potential grammatical inconsistencies in the manuscript.

In addition, our evaluation pipeline relies on API-based access to multiple commercial and open-source AI systems. We used these APIs to compare model behaviors and capabilities, and to assess the reliability, difficulty, and discriminative power of \textsc{RealClawBench}. These runs were integral to validating whether the benchmark distinguishes performance differences among diverse AI systems.

AI-assisted coding tools were also used during implementation and experimentation to help debug scripts, inspect runtime errors, analyze logs, and improve engineering efficiency. Furthermore, some figures and visual illustrations in the paper were generated or refined with the assistance of AI-based image generation and visualization tools to improve presentation quality and readability.

All AI-generated or AI-assisted content, including text revisions, code suggestions, evaluation runs, and figures, was carefully reviewed, independently verified, and edited by the authors before inclusion in the final manuscript.

\end{document}